\documentclass[a4paper, 10pt, conference]{ieeeconf}      

\IEEEoverridecommandlockouts                              

\usepackage{graphicx}
\graphicspath{{figs/}}
\usepackage{amsmath}
\usepackage{amssymb}
\usepackage{algorithm}
\usepackage{algpseudocode}
\usepackage{times}
\usepackage{subfigure}
\newcommand{\figref}[1]{Fig.~\ref{figure:#1}}

\newcommand{\secref}[1]{Section \ref{sec:#1}}

\title{\LARGE \bf
 Front Hair Styling Robot System Using Path Planning for Root-Centric Strand Adjustment
}

\author{Soonhyo Kim$^{1}$,
        Naoaki Kanazawa$^{1}$,
        Shun Hasegawa$^{1}$,
        Kento Kawaharazuka$^{1}$
        and Kei Okada$^{1}$ 
        \thanks{This work was partially supported by JST SPRING, Grant Number JPMJSP2108,
          $^{1}$S. Kim, N. Kanazawa, S. Hasegawa, K. Kwaharazuka and K. Okada are with JSK Laboratory, Graduate School of Information Science and Technology,
        the University of Tokyo, 7-3-1, Hongo, Bunkyo-ku, Tokyo, Japan
        }
}

\begin{document}

\maketitle
\thispagestyle{empty}
\pagestyle{empty}

\begin{abstract}
  Hair styling is a crucial aspect of personal grooming, significantly influenced by the appearance of front hair. While brushing is commonly used both to detangle hair and for styling purposes, existing research primarily focuses on robotic systems for detangling hair, with limited exploration into robotic hair styling. This research presents a novel robotic system designed to automatically adjust front hairstyles, with an emphasis on path planning for root-centric strand adjustment. The system utilizes images to compare the current hair state with the desired target state through an orientation map of hair strands. By concentrating on the differences in hair orientation and specifically targeting adjustments at the root of each strand, the system performs detailed styling tasks. The path planning approach ensures effective alignment of the hairstyle with the target, and a closed-loop mechanism refines these adjustments to accurately evolve the hairstyle towards the desired outcome. Experimental results demonstrate that the proposed system achieves a high degree of similarity and consistency in front hair styling, showing promising results for automated, precise hairstyle adjustments.

\end{abstract}

\section{Introduction}
Hair styling is an essential aspect of personal grooming and self-expression. While there has been significant progress in automating various grooming tasks, the automation of hair styling, particularly with a robotic system, remains an underexplored area. The advent of image-based technologies opens up new possibilities for automating such tasks with increased flexibility.

This research presents a novel system that utilizes target images as a basis for hair styling, specifically focusing on the front hair. The system leverages an image-based goal-setting framework, allowing it to adapt to various target hairstyles derived from in-the-wild photos. This method offers a flexible, autonomous solution for hairstyling by continuously adjusting hair based on comparisons between the current state and the target image. By using path planning for a root-centric strand adjustment approach , the system generates effective hair stroke motions, resulting in styles that closely resemble the target. Experimental results quantitatively evaluate the system's ability to achieve the desired style, demonstrating its effectiveness and versatility.

Moreover, the image-based nature of this system enables seamless integration with various hair editing applications, providing significant scalability and a broader range of potential use cases.

\begin{figure}[!t]
 \begin{center}
  \includegraphics[width=\columnwidth]{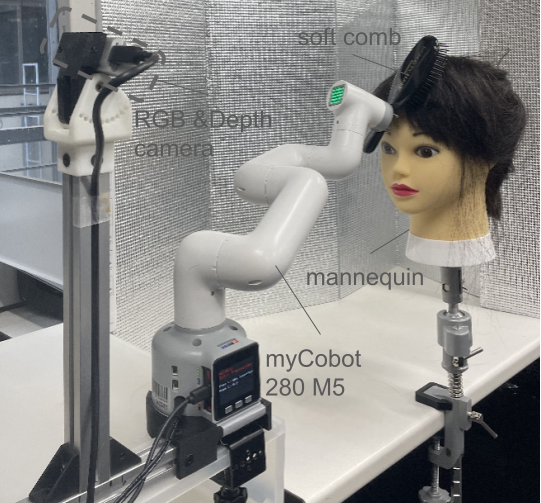}
  \caption{System setup for front hair styling robot system}
  \label{figure:intro}
 \end{center}
\end{figure}

\section{Related Works}

\subsection{Robotic Hair manipulation}
Robotic hair styling has not been extensively explored in robotics. Some earlier works have investigated robotic systems for brushing or manipulating hair, including force control methods \cite{kuka}, feedback-driven hair detangling motions \cite{detangling_hair}, and the introduction of soft robotic end-effector designs for contact-rich hair manipulation \cite{moe}. However, these systems generally do not address the complexities of dynamic styling tasks that involve adjustments based on visual feedback or target appearance. Furthermore, there has been no prior approach that directly utilizes Root-Centric Strand Adjustment for hairstyle modification.

Existing methods, while incorporating image-based techniques such as extracting hair strands from orientation maps and employing them for stroke-based path planning\cite{hair_combing_system}, have not explored the unique advantages offered by a Root-Centric Strand Adjustment method. Our approach leverages the corresponding root positions of specific hair strands with target hair strands during the path planning process, allowing for more effective stroke generation. By incorporating the current hair's root into the stroke path, the system is able to plan more controlled combing strokes that better replicate the target hairstyle. This root-centric strategy not only enhances the overall control hair strands and adaptability of the styling process but also ensures that the transformation closely aligns with the target hairstyle. Our system offers a more efficient and accurate way to achieve desired hairstyles through the effective planning and execution of root-centric stroke paths.

\subsection{Target-Based Goal Setting in Robotics}

In robotics, target images have been effectively used for goal specification in various tasks. A lot of rigid object alignment tasks is conducted from using the target image\cite{flow-based, demonstration, dall-e-bot, BAIR}. Also, in deformable object manipulation\cite{localgnn, cloth, without_demo}, the target image is utilized as a object representation.

While target-based approaches are often established in other domains, applying them to robotic hair styling is novel. The ability to adapt the system based on target images allows for the flexible execution of various hair styling tasks. This flexibility is further enhanced by the system's compatibility with in-the-wild photos, making it adaptable to a broad range of hairstyles.

\section{Methodology}

This section outlines the methodology developed for automatic front hair styling, which iteratively adjusts the current hair state to a target hairstyle. The system follows a modular pipeline with stages for hair orientation comparison, path generation, and trajectory creation, designed to optimize robotic hair manipulation. The methodology ensures that the hairstyle evolves towards the target appearance through iterative refinement.

\subsection{Overview}

\begin{figure*}[t!]
    \begin{center}
        \includegraphics[width=2.0\columnwidth]{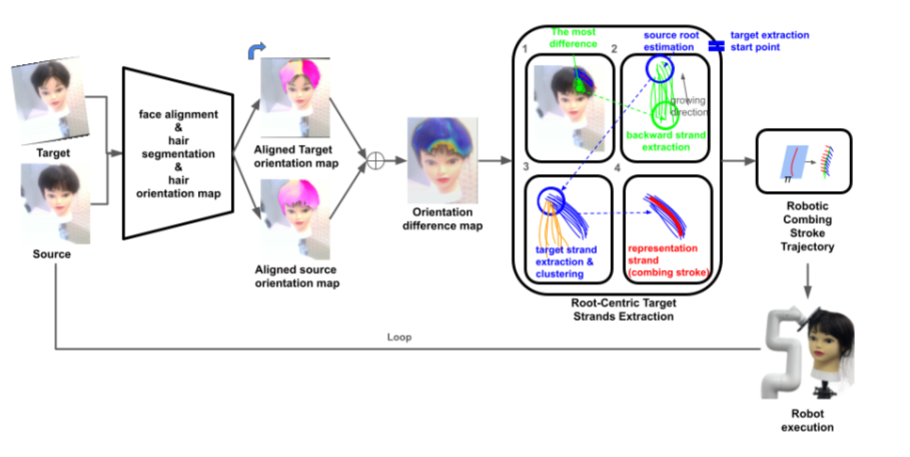}
        \caption{System overview for front hair styling robot system}
        \label{figure:overview}
    \end{center}
\end{figure*}

\figref{overview} illustrates the overall pipeline of the proposed system. Each stage is designed to work in sequence, starting from comparing the current and target hair states, extracting relevant hair strands, and generating combing strokes to adjust the hairstyle. The final stage involves converting these strokes into 3D trajectories for robotic execution, ensuring that the hairstyle evolves towards the target appearance through iterative adjustments.

The proposed pipeline seeks to compute the differences between the current and target hairstyles and generate feasible trajectories for robotic manipulation. The system operates in three primary stages:

\begin{enumerate}
    \item \textbf{Target Hair Set Comparison}: Analyzing the differences between the current hair and the target hairstyle.
    \item \textbf{Combing Stroke Generation from Root-Centric Strand Adjustment}: Generating combing strokes from extracting corresponding root-centric target strands for manipulating current hair strands based on identified differences.
    \item \textbf{Robotic Combing Stroke Trajectory}: Converting 2D manipulation strokes into 3D trajectories for robotic execution.
\end{enumerate}

Each step of the process is designed to iteratively refine the front hairstyle, ensuring that the hairstyle becomes more similar to the desired target appearance.

\subsection{Target Hair Set Comparison}

This phase focuses on comparing the current hair state with the target hairstyle. The system begins by aligning the target hairstyle with the subject’s facial structure and calculating the differences in hair orientation. This process is crucial for ensuring that the robotic system can recognize and correct the deviations between the current and target hairstyle.

\subsubsection{Target Hair Set Representation}
The target hairstyle is provided as a reference, \( I_{\text{target}} \), which is a real image captured in the wild. The system aligns \( I_{\text{target}} \) with the subject’s facial geometry using face alignment techniques such as Mediapipe~\cite{mp}, which ensures proper orientation and placement.

Let \( I_{\text{current}} \) represent the current hairstyle, and \( I_{\text{aligned}} \) be the aligned target image. The alignment function \( f_{\text{align}} \) maps the target hairstyle onto the current face shape:

\begin{equation}
I_{\text{aligned}} = f_{\text{align}}(I_{\text{target}}, I_{\text{current}}).
\end{equation}

\subsubsection{Hair Segmentation and Orientation Extraction}
Using a segmentation mask \( M_{\text{hair}} \), the hair region is isolated. To reduce edge noise, a morphological erosion process is applied with a structuring element \( B \) (in this research, we used size 15):

\begin{equation}
M_{\text{hair}}' = M_{\text{hair}} \ominus B,
\end{equation}

This cleaned mask allows for the extraction of an orientation map \( \theta \), which represents the direction of hair strands. We used the HairStep~\cite{hairstep} deep learning-based orientation map for precise strand direction extraction.

\subsubsection{Difference Calculation}
The system computes the orientation difference between the current and aligned target hairstyles over the current mask region. Let \( \theta_{\text{current}} \) and \( \theta_{\text{aligned}} \) denote the orientation maps for the current and target hairstyles, respectively. The difference map \( \Delta_{\theta} \) is computed as:

\begin{equation}
\Delta_{\theta} = |\theta_{\text{aligned}} - \theta_{\text{current}}|.
\end{equation}

If \( \Delta_{\theta} > 180^\circ \), the value is adjusted by subtracting 180°, ensuring that the difference remains within the range of 0° to 180°.

In this system, the comparison includes areas where one of the images may not have a hair mask. Specifically, the orientation difference is computed even if one mask is present and the other is absent, allowing the system to capture discrepancies in uncovered hair regions. The overall deviation is quantified using the average orientation difference over the entire current mask region \( M_{\text{current}} \):

\begin{equation}
\bar{\Delta}_{\theta} = \frac{1}{|M_{\text{current}}|} \sum_{(x,y) \in M_{\text{current}}} |\Delta_{\theta}(x,y)|.
\end{equation}

\subsection{Stroke Generation from Root-Centric Strands Extraction}

Based on the computed orientation differences, the system identifies and manipulates the current hair strands in regions requiring adjustment. This process involves strand extraction, sampling, and generating smooth manipulation paths. This step is vital for ensuring smooth transitions between the current hairstyle and the target.

\subsubsection{Hair Strand Extraction Near Differences}
The system identifies hair strands in regions where \( \Delta_{\theta} \) is significant. For each extracted strand \( S_{\text{current}} \), the root point is determined and denoted as \( p_{\text{root}} \).

\subsubsection{Root-Centric Target Strands Extraction}

To estimate the corresponding root-centric target strands for adjustment with current strands, we transform each sampled target strand’s end point and the direction vector (calculated as the difference between the end point and the root) into a feature space. This feature space represents both the spatial position and orientation of the strands. We apply the DBSCAN (Density-Based Spatial Clustering of Applications with Noise) \cite{dbscan} algorithm to group these target strands into clusters based on similar directional properties.

By clustering the strands in this way, we can segment the hair into bundles with similar orientation. The cluster with the largest size is selected as it represents the dominant directionality within that region. The representative strand for manipulation is then computed by averaging the directional vectors of all strands within the largest cluster. This averaged strand provides a smooth, root-aligned reference strand \( S_{\text{rep}} \), which is used to generate combing strokes:

\begin{equation}
S_{\text{rep}} = \frac{1}{|C_{\text{max}}|} \sum_{i \in C_{\text{max}}} S_{\text{target},i},
\end{equation}

where \( C_{\text{max}} \) is the cluster with the largest number of strands. The representative strand \( S_{\text{rep}} \) is used to guide the robotic combing stroke generation.

\subsubsection{Current Hair Strand Manipulation}
The extracted current strands are adjusted based on the length of the comb used. For a strand \( S_{\text{current}} \) with length \( L_{\text{current}} \), the adjustment accounts for the comb length \( L_{\text{comb}} \). The root point \( p_{\text{root}} \) is shifted by \( \Delta L = L_{\text{comb}} \), modifying the length as follows:

\begin{equation}
L_{\text{adjusted}} = L_{\text{current}} + \Delta L.
\end{equation}

To ensure smoothness, a Bezier curve \cite{bezier} \( f_{\text{smooth}} \) is applied to the manipulated strand:

\begin{equation}
S_{\text{smoothed}} = f_{\text{smooth}}(S_{\text{current}}).
\end{equation}

\subsection{Robotic Combing Stroke Trajectory}

In this phase, we convert 2D manipulation strokes into 3D trajectories for real robotic execution.

\subsubsection{Cartesian Positioning}
Depth images are used to convert 2D strokes into Cartesian space, accurately defining their positions in 3D.

\subsubsection{Trajectory Pose Calculation}
RANSAC (Random Sample Consensus) \cite{ransac} is employed to estimate the orientation of the z-axis for the robot's end-effector. By analyzing a set of points \( P = \{p_1, p_2, \dots, p_n\} \) around the 2D stroke, RANSAC determines the plane \( \pi \) and defines the z-axis direction:

\begin{equation}
\pi : ax + by + cz + d = 0.
\end{equation}

The system projects the strokes onto the estimated plane \( \pi \), setting the y-axis based on the path direction, and calculates the x-axis as the cross product of the y-axis and z-axis:

\begin{equation}
x = y \times z.
\end{equation}

The 2D strokes are then extruded into 3D space to generate a complete 3D trajectory for the robotic system.

\section{Experiments}
\label{sec:experiments}

\subsection{Experiment Objective}

The primary goal of the experiments is to validate the accuracy and effectiveness of the proposed Path Planning for Root-Centric Strand Adjustment system for front hairstyling tasks, specifically assessing its capability to reproduce target hairstyles by dynamically adjusting hair strand orientations. The system is compared against a random strand extraction methods to measure improvements in styling accuracy.

\subsection{Experiment Setup}

The experiments involved comparing two methods:

\begin{itemize}
    \item \textbf{Path Planning for Root-Centric Strand Adjustment}: A system that adjusts hair configuration by planning paths rooted at each strand's base and tracing the corresponding target strand to match the desired hairstyle.
    \item \textbf{Random Strand Extraction Methods}: A method where strokes for target hair strands are chosen randomly, without the benefit of a root-centric path planning mechanism.
\end{itemize}

For the experimental setup, we used an RGB-depth camera, a myCobot 280 M5\cite{mycobot_280_m5} attached with a soft comb, and a mannequin head with stiff hair. To facilitate manipulation, water was sprayed onto the mannequin's hair before each trial. At the beginning of each experiment, the hair was brushed down naturally from top to bottom to ensure consistent initial conditions.

\subsection{Evaluation Metrics}

The system's performance was evaluated using the following metric:

\begin{itemize}
    \item \textbf{Styling Accuracy}: Measured by orientation differences (\(\Delta_{\theta}\)) between the final styled hair and the target hairstyle. Lower values indicate higher accuracy in matching the target hairstyle.
\end{itemize}

\subsection{Results}

The proposed Path Planning for Root-Centric Strand Adjustment system demonstrated superior performance compared to the random strand extraction methods in terms of accuracy. The results are summarized in Table \ref{table:multi_target_results}.

\begin{figure}[!t]
 \begin{center}
  \includegraphics[width=\columnwidth]{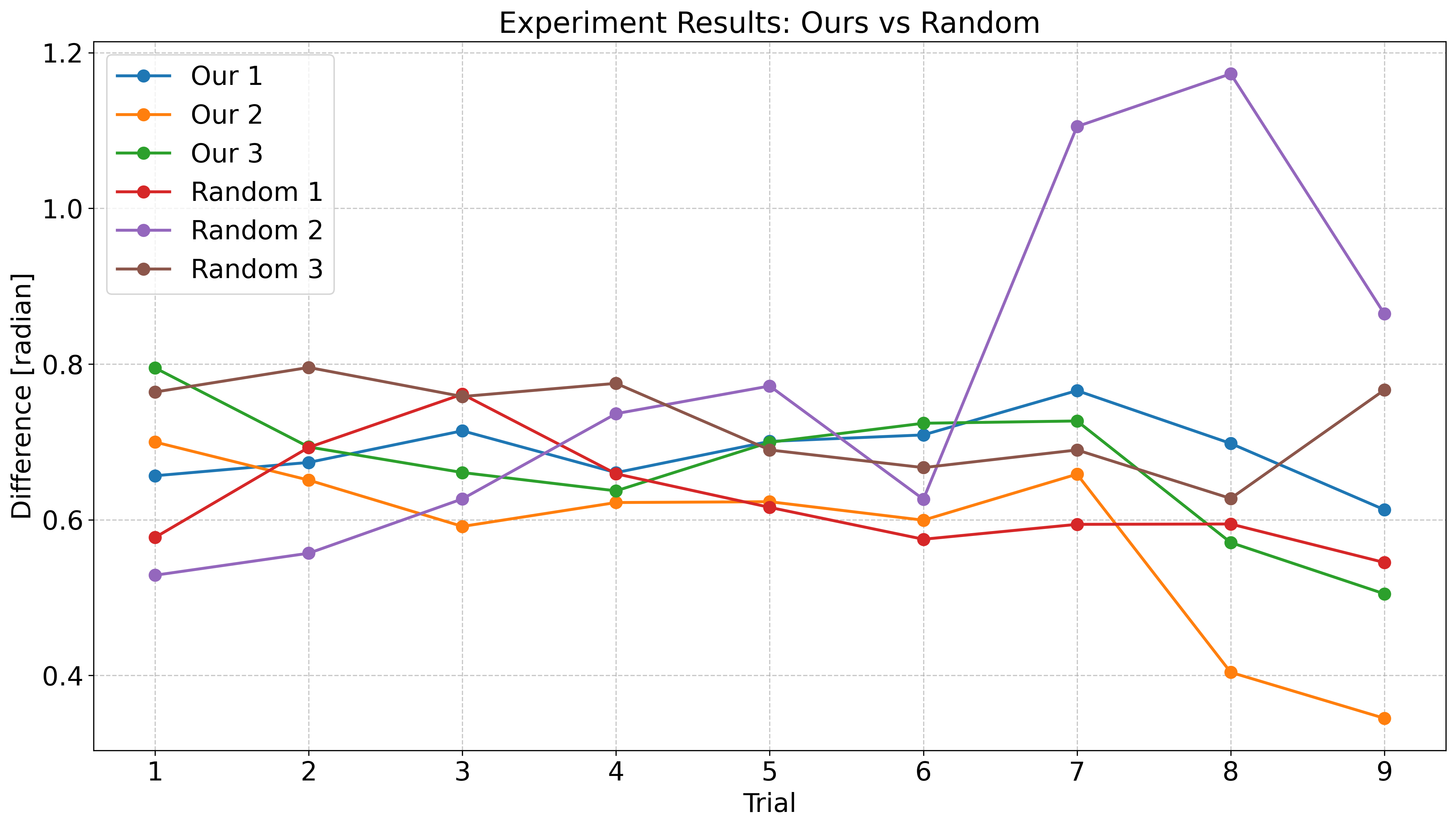}
  \caption{Orientation difference over iterations for our proposed system and the random sampling method.}
  \label{figure:experiment_result}
 \end{center}
\end{figure}

\begin{figure}[!t]
 \begin{center}
  \includegraphics[width=\columnwidth]{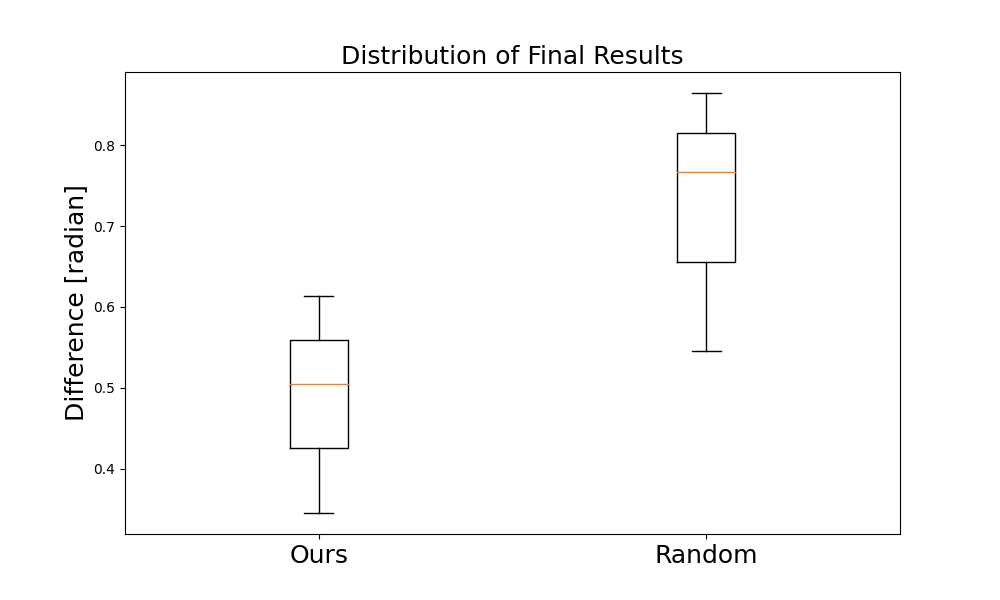}
  \caption{Box plot comparison of orientation differences between our proposed system and the random sampling method.}
  \label{figure:boxplot}
 \end{center}
\end{figure}

\begin{table}[htb]
\centering
\caption{Performance Comparison between Proposed System and Random Baseline}
\begin{tabular}{|c|c|c|}
\hline
\textbf{Method} & \textbf{Mean Difference (rad)} & \textbf{Standard Deviation (rad)} \\ \hline
Ours & 0.4877 & 0.1101 \\ \hline
Random & 0.7255 & 0.1336 \\ \hline
\end{tabular}
\label{table:multi_target_results}
\end{table}

As shown in Table \ref{table:multi_target_results}, our system achieved a mean orientation difference of 0.4877 radians with a standard deviation of 0.1101 radians, compared to the random sampling method's mean difference of 0.7255 radians and standard deviation of 0.1336 radians. These results indicate that the proposed root-centric strand adjustment approach consistently outperforms the random baseline in achieving the desired hairstyle, with better accuracy and lower variability in outcomes.

\figref{experiment_result} illustrates the change in orientation difference over iterations for both methods, showing that our system converges more quickly and consistently toward the target style, while the random sampling method exhibits more erratic performance.

The box plot in \figref{boxplot} further highlights the superior performance of our system, with a lower median and smaller interquartile range, indicating more consistent and accurate results compared to the random approach.

\figref{result_5to5} provides a visual presentation of the final results:

\begin{figure*}[htb!]
 \begin{center}
   \includegraphics[width=\textwidth]{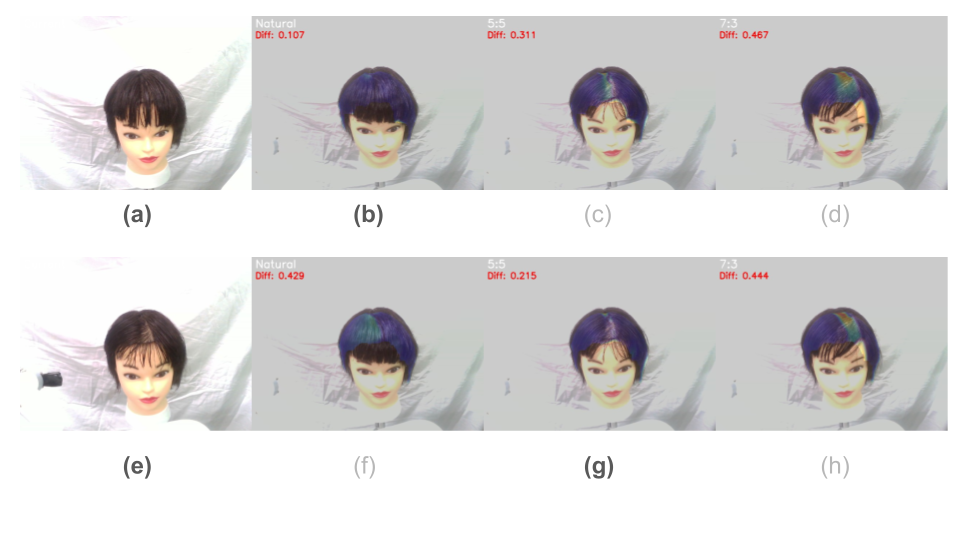}
   \caption{
     Final hair styling results. The image shows, from left to right:
     \textbf{(a)} the initial natural hair state before styling,
     \textbf{(b)} comparison between the initial state and a natural hairstyle,
     (c) comparison between the initial state and a 5:5 parting hairstyle,
     (d) comparison between the initial state and a 7:3 parting hairstyle,
     \textbf{(e)} the final result produced by our system,
     (f) comparison between the final result and a natural hairstyle,
     \textbf{(g)} comparison between the final result and a 5:5 parting hairstyle, and
     (h) comparison between the final result and a 7:3 parting hairstyle.
     Numerical values, indicated in red text, represent orientation differences, where smaller values indicate higher similarity to the target style.
     Notably, \textbf{(b)} has the smallest difference value, representing the closest similarity to the natural style, and \textbf{(g)} has the smallest difference value among the styled results, confirming the system's success in achieving the target 5:5 parting hairstyle.
   }
   \label{figure:result_5to5}
 \end{center}
\end{figure*}

The image in \figref{result_5to5} illustrates the results of our system for automatic front hair styling, with the target hairstyle being a 5:5 parting. The comparisons are as follows:

- \textbf{(b), (c), and (d)}: These show comparisons between the initial natural hair state and each target style (natural, 5:5 parting, and 7:3 parting). These provide a baseline for assessing the system's performance. Notably, the initial state \textbf{(a)} closely resembles the natural hairstyle \textbf{(b)}, as evident from the similarity in scores and visual appearance. \textbf{(b)} also has the smallest difference value among the initial comparisons.

- \textbf{(f), (g), and (h)}: These show comparisons between the final hairstyle produced by the system \textbf{(e)} and the natural hairstyle, the 5:5 parting hairstyle, and the 7:3 parting hairstyle, respectively. The final result \textbf{(e)} closely matches the target 5:5 parting hairstyle \textbf{(g)}, as demonstrated by the low difference value (indicated in red) and visual similarity. \textbf{(g)} has the smallest difference value among the styled results, confirming the system's effectiveness in achieving the desired style.

The heatmap in the figure, overlaid on each comparison, visually represents orientation differences. Blue regions indicate higher similarity, while red regions highlight larger directional discrepancies. The scores, displayed in red text, represent the average value of the corresponding heatmap. These scores, combined with the heatmap and visual comparisons, confirm that the system effectively achieved the desired 5:5 parting style. The root-centric path planning approach ensured consistent and accurate results, adapting the initial state into the desired target style while minimizing orientation differences.

The results demonstrate the system's ability to produce hairstyles that closely align with the target, highlighting its effectiveness in both numerical evaluation and visual appearance.

\section{Conclusion}
Our hair Root-Centric Strand Adjustment System demonstrates significant advantages over the random sampling baseline in front hairstyling tasks. The 32.8\% reduction in mean orientation difference highlights the system's superior accuracy in replicating target hairstyles. Additionally, the lower standard deviation reflects improved consistency across trials, which is crucial for automated grooming applications.

As shown in \secref{experiments}, our system maintains stability across iterations, in contrast to the erratic behavior observed with the random method. This suggests that the root-centric path planning mechanism effectively guides the styling process and minimizes errors.

Furthermore, \secref{experiments} illustrates the practical benefits of our system, producing more natural and aesthetically pleasing hairstyles. The ability to closely match the target style is essential in hairstyling, where even subtle differences can significantly affect the overall appearance. The visual and numerical comparisons in the results confirm that the system successfully achieves a hairstyle that is highly similar to the target.

However, some limitations remain in our system:

(i) Brushing intensity based on responsiveness:
Brushing intensity is determined by the comb's posture calculated by RANSAC and a manually fixed brush size in our system. However, the system lacks the ability to adapt to real-time feedback, such as increasing the brushing force or dynamically adjusting the brushing direction based on hair response. Future improvements should focus on enabling the system to adapt its behavior in response to hair feedback.

(ii) Path planning considering the 3D geometric structure of the hair:
In this research, the system converts a 2D path into a 3D trajectory using depth images from a single view. This approach limits path generation for hairstyles like slicked-back hair or other areas that are not directly visible. More advanced path planning that takes into account the full 3D geometry of the hair is needed.

(iii) Path planning considering the comb size:
The current system uses a fixed comb size to segment the path. More flexible path planning that accounts for the comb’s reach could prevent interference with hair strands oriented in different directions. A hierarchical path planning approach, which considers the complexity of the hairstyle, could improve the handling of intricate hairstyles.

(iv) Limitations of brushing alone:
There are inherent limitations to the types of hairstyles that can be achieved solely through brushing. The system cannot adapt to more complex styles that require additional actions, such as applying water to increase hair pliability. Integrating auxiliary tools could expand the range of possible hairstyles, making the system more versatile in handling diverse hair types and styling needs.

\addtolength{\textheight}{-12cm}   

\bibliographystyle{junsrt}
\bibliography{main}

\end{document}